\documentclass{article}

\usepackage[preprint]{icml2026}

\usepackage[utf8]{inputenc} %
\usepackage[T1]{fontenc}    %
\usepackage{hyperref}       %
\usepackage{url}            %
\usepackage{booktabs}       %
\usepackage{amsfonts}       %
\usepackage{nicefrac}       %
\usepackage{microtype}      %
\usepackage{xcolor}         %
\usepackage{graphicx}
\usepackage{enumitem}
\usepackage{amsmath}
\usepackage{subcaption}
\usepackage{xspace}
\usepackage{letterspace}

\usepackage{tikz}
\usetikzlibrary{backgrounds,calc,positioning,arrows.meta,shadows,fit}
\usepackage{cleveref}

\icmltitlerunning{Position: Capability Control Should be a Separate Goal From Alignment}

\begin{document}

\twocolumn[
  \icmltitle{Position: Capability Control Should be a Separate Goal From Alignment}

  \icmlsetsymbol{equal}{*}

  \begin{icmlauthorlist}
    \icmlauthor{Shoaib Ahmed Siddiqui}{cam}
    \icmlauthor{Eleni Triantafillou}{gdm}
    \icmlauthor{David Krueger}{uom,mila}
    \icmlauthor{Adrian Weller}{cam,alan}
  \end{icmlauthorlist}

  \icmlaffiliation{cam}{University of Cambridge}
  \icmlaffiliation{gdm}{Google DeepMind}
  \icmlaffiliation{uom}{University of Montreal}
  \icmlaffiliation{mila}{Mila}
  \icmlaffiliation{alan}{The Alan Turing Institute}
  \icmlcorrespondingauthor{Shoaib Ahmed Siddiqui}{msas3@cam.ac.uk}
  \icmlcorrespondingauthor{Eleni Triantafillou}{etriantafillou@google.com}

  \icmlkeywords{Capability Control, Safety, Alignment, RLHF, Unlearning, Steering.}

  \vskip 0.3in
]

\printAffiliationsAndNotice{}  %

\begin{abstract}
Foundation models are trained on broad data distributions, yielding generalist capabilities that enable many downstream applications but also expand the space of potential misuse and failures.
This position paper argues that \emph{capability control}---imposing restrictions on permissible model behavior---should be treated as a distinct goal from alignment.
While alignment is often context and preference-driven, capability control aims to impose hard operational limits on permissible behaviors, including under adversarial elicitation.
We organize capability control mechanisms across the model lifecycle into three layers: (i) \emph{data-based} control of the training distribution, (ii) \emph{learning-based} control via weight- or representation-level interventions, and (iii) \emph{system-based} control via post-deployment guardrails over inputs, outputs, and actions.
Because each layer has characteristic failure modes when used in isolation, we advocate for a defense-in-depth approach that composes complementary controls across the full stack.
We further outline key open challenges in achieving such control, including the dual-use nature of knowledge and compositional generalization.
\end{abstract}

\section{Introduction}
\label{sec:intro}

Recent progress in deep learning has produced breakthroughs in a wide range of domains, including game-playing \citep{mnih2013atari,silver2017masteringgo,berner2019dota}, weather forecasting \citep{pathak2022fourcastnet,lam2023learning,bi2023pangu}, protein folding \citep{jumper2021highly}, algorithmic discovery \citep{novikov2025alphaevolve}, and autonomous driving \citep{bojarski2016end}.
Historically, these systems were task-specific---i.e., meticulously designed to excel at a single objective---often referred to as \textit{narrow AI} \citep{michaud2025creation}.
However, the field has recently shifted towards training generalist systems, widely referred to as \textit{Foundation Models} \citep{bommasani2021foundationmodels}.
These models serve as a robust basis for a multitude of different downstream applications.

Training these foundation models has resulted in simultaneous improvements across a broad spectrum of tasks, alongside the emergence of novel capabilities \citep{wei2022emergent}.
A notable example among such capabilities is \textit{in-context learning}---i.e., the ability to learn new tasks solely from examples provided within the context window \citep{lu2023emergent}.
Despite their utility, foundation models pose unique challenges, particularly associated with the security and privacy of such systems.

Even task-specific systems, which are constrained by design, have been shown to exhibit undesirable and exploitable behaviors \citep{szegedy2013intriguing,alcorn2019strike,ollikka2024comparison,gu2019badnets,shokri2017membership}.
These risks are further exacerbated for open-domain foundation models, which are trained to be versatile and general in order to serve as a strong foundation for a variety of downstream tasks \citep{bommasani2021foundationmodels}.
Effective capability control is therefore essential to minimize risks in real-world deployments, guarding against both malicious misuse by bad actors and inadvertent failure modes \citep{qi2023finetuning,li2024wmdp}.
Large Language Models (LLMs)---a canonical example of foundation models---typically undergo a rigorous post-training phase to suppress such dangerous capabilities \citep{bai2022constitutional,touvron2023llama,team2024gemma}.

\begin{figure*}[t]
    \centering
    \resizebox{\textwidth}{!}{  %
    \begin{tikzpicture}[
        node distance=1.2cm and 1.8cm, %
        every node/.style={font=\sffamily},
        block/.style={
            rectangle,
            draw,
            thick,
            rounded corners,
            minimum height=1.2cm,
            text width=3.0cm,
            align=center,
            inner sep=6pt,
            drop shadow %
        },
        learningblock/.style={block, fill=green!10, draw=green!50!black},
        modelblock/.style={block, fill=orange!10, draw=orange!60!black},
        arrow/.style={thick,->,>=stealth, color=gray!80!black},
        edgeLabel/.style={font=\footnotesize\sffamily\bfseries, color=gray!70!black, align=center, midway, above=2pt},
        controlregion/.style={draw, dashed, thick, rounded corners, inner sep=10pt},
        databg/.style={controlregion, draw=blue!40!gray, fill=blue!5},
        learnbg/.style={controlregion, draw=green!40!black, fill=green!4},
        systembg/.style={controlregion, draw=red!50!black, fill=red!4}
    ]

    \node[block, fill=blue!7, draw=blue!50!black] (data) {
        \textbf{Training Data}\\[3pt]
        \footnotesize Large-scale corpus
    };

    \node[learningblock, right=2.0cm of data] (pretrain) {
        \textbf{Pretraining}\\[3pt]
        \footnotesize Learn broad capabilities
    };

    \node[learningblock, right=2.2cm of pretrain] (alignment) {
        \textbf{Alignment}\\[3pt]
        \footnotesize Align to (collective) human values
    };

    \node[modelblock, right=2.2cm of alignment, text width=3.2cm] (system) {
        \textbf{Deployed System}\\[3pt]
        \footnotesize Foundation model within tools / agents / APIs
    };

    \draw[arrow] (data) -- node[edgeLabel] {Training\\Distribution} (pretrain);

    \draw[arrow] (pretrain) -- node[edgeLabel] {Pretrained\\Model} (alignment);

    \draw[arrow] (alignment) -- node[edgeLabel] {Aligned\\Model} (system);

    \begin{scope}[on background layer]
        \node[databg, fit=(data)] (datacontrol) {};
        \node[above=0.15cm of datacontrol.north, font=\bfseries\footnotesize, color=blue!60!black]
            {Data-based Capability Control};
        \node[below=0.05cm of datacontrol.south, font=\scriptsize, color=blue!60!black]
            {Filter / curate / synthesize training data};

        \node[learnbg, fit=(pretrain)(alignment)] (learncontrol) {};
        \node[above=0.15cm of learncontrol.north, font=\bfseries\footnotesize, color=green!50!black]
            {Learning-based Capability Control};
        \node[below=0.05cm of learncontrol.south, font=\scriptsize, color=green!50!black]
            {SFT, RLHF, unlearning, adversarial training (at pretraining or alignment)};

        \node[systembg, fit=(system)] (syscontrol) {};
        \node[above=0.15cm of syscontrol.north, font=\bfseries\footnotesize, color=red!70!black]
            {System-based Capability Control};
        \node[below=0.05cm of syscontrol.south, font=\scriptsize, color=red!70!black]
            {Input/output filters, CoT monitors, information-flow policies};
    \end{scope}

    \end{tikzpicture}
    }
    \caption{\textbf{Overview of Capability Control Layers.} Capability control can be applied at three complementary levels across the model lifecycle: (i) \textit{data-based} control shapes the training distribution of foundation models; (ii) \textit{learning-based} control modifies model weights or representations during training; and (iii) \textit{system-based} control constrains the behavior of the deployed system via guardrails on inputs, outputs, and tool access.}
    \label{fig:alignment-vs-cap}
\end{figure*}
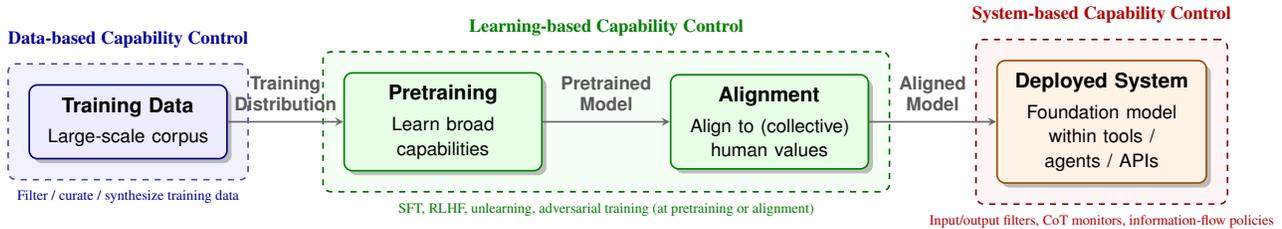

Capability control is currently viewed as an implicit part of the alignment process \citep{bai2022constitutional,edelman2025full,touvron2023llama}.
\textbf{This position paper argues that capability control should be treated as a separate goal rather than merely a component of the alignment pipeline}.
To this end, we must establish the levels of abstraction at which control can be effectively enforced.
As illustrated in \Cref{fig:alignment-vs-cap}, the model lifecycle progresses from large-scale data to a trained model, and finally to a deployed system.
Consequently, capability control can be enforced at three distinct levels: (i) controlling the training distribution, (ii) modifying internal weights or representations, or (iii) enforcing guardrails at the system level.
While the subsequent sections explore these levels in detail, we argue that none is sufficient in isolation due to their inherent limitations.
Therefore, we advocate for a \textit{defense-in-depth} approach that imposes complementary interventions across the entire stack (\Cref{sec:defense_in_depth}).

\section{Capability Control}

\subsection{Objective of Capability Control}

Capability control is the set of methodological interventions applied to a foundation model---at the data, learning, or system level---aimed at defining and enforcing operational boundaries on permissible model behaviors.
Specifically, capability control entails the inhibition or erasure of specific functional capacities (e.g., hazardous knowledge, unsafe tool usage) such that they cannot be elicited in either benign or adversarial use cases.

\subsection{Realization of Capability Control}

We discuss approaches to realize capability control across the model's lifecycle, forming a \textit{defense-in-depth} mechanism.
These interventions are broadly categorized into: (i) \textit{data-based control}, which focuses on data-based interventions for control;
(ii) \textit{learning-based control}, which formulates capability control as a learning problem; and
(iii) \textit{System-based control}, which enforces control at inference time via external guardrails.

While we consider absolute prohibition of harmful or non-permissible capabilities as the objective, the operationalization of these mechanisms is often imperfect in practice.
This arises from the inherent probabilistic nature of current learning paradigms, the dual-use nature of knowledge, and the susceptibility of guardrails to adversarial circumvention.
We discuss these practical limitations and open challenges in detail in \Cref{sec:open_challenges}. Nonetheless, we believe that significant progress can be made, with meaningful consequences for real-world problems, even without perfectly achieving the goal.

\subsection{Does Alignment Cover Capability Control?}
\label{subsec:diff_alignment_cap}

Most modern post-training pipelines simultaneously optimize for both alignment and capability control \citep{touvron2023llama,team2024gemma,yang2025qwen3,liu2024deepseek,bai2022constitutional}.
This raises the question: \textit{does alignment implicitly cover capability control?}

As alignment aims to make models follow (collective) human intent and values, it is often heavily dependent on the context---e.g., things that might be appropriate in an educational context might turn out to be unethical or even illegal in other cases \citep{ngo2022alignment,ouyang2022instructgpt,longpre2023flan,korbak2024aligning,edelman2025full}.
This context dependence has been widely exploited as an attack vector, where harmful requests are reframed as educational prompts \citep{jailbreaking_chatgpt_release_day,andriushchenko2024does}.

By contrast, we define capability control as specifying hard limits on what the system is allowed to do\footnote{Such hard limits on capabilities have also been defined as red-lines in the past \citep{WEF_2025_AI_redlines}.}.
In this sense, alignment can be ``soft'' and context-sensitive, whereas capability control aims to impose \emph{absolute} prohibitions on certain behaviors, regardless of context.
Under our notion of capability control, detailed operational guidance on, e.g., enhancing human pathogens to increase their lethality would be disallowed \emph{in any context}; such a capability is treated as outside the model's permissible behavior altogether.

If one treats capability control as merely a byproduct of (post-)training alignment, one may primarily focus on training-time interventions \citep{korbak2023pretraining,touvron2023llama,team2024gemma,claudeconstitution}, as visualized in \Cref{fig:alignment-vs-cap}.
However, we will argue in this paper that capability control can and should be exercised at all levels within this cycle, even after model deployment via system-level guardrails.
Note that despite this conceptual distinction, many of the methods employed for alignment can be readily adapted as effective tools for capability control.

\textbf{Alignment Tax.}
It is widely recognized that alignment incurs a ``tax'' in terms of development time, computational resources, and performance degradation on downstream tasks \citep{alignment_forum_alignment_tax,lin2024alignmenttax}.
However, because the alignment and capability control phases are deeply entangled in modern post-training pipelines \citep{touvron2023llama,team2024gemma}, isolating the source of this degradation remains difficult.

By treating alignment and capability control as distinct processes, we conjecture that the majority of the alignment tax---particularly performance degradation---is driven by capability control rather than the alignment process itself \citep{ouyang2022instructgpt,longpre2023flan}.
Observable failure modes such as benign over-refusal \citep{cui2024or} support this hypothesis, as they manifest directly from safety guardrails rather than the model's fundamental ability to follow user intent.

\subsection{Difference between Capability Suppression and Removal}

It is critical to distinguish capability suppression from removal.
While the ultimate aim is often to permanently erase undesired capabilities, thereby guaranteeing non-elicitability, many existing methods merely suppress them \citep{ren2025sok,siddiqui2025dormant}.
System-based interventions, for instance, offer only conditional mediation by design.
Similarly, standard learning-based methods---particularly those relying on behavior demonstrations or preference data---have been shown to primarily suppress capabilities rather than remove them \citep{zou2023universal,che2025tamperingattacks}.
In contrast, specific data-centric and unlearning-based approaches generally promise actual knowledge removal, ensuring a capability cannot be elicited \citep{o2025deepignorance,maini2025safety,tamirisa2408tamper}.
However, even this premise has been recently challenged, with new evidence suggesting that unlearning techniques may also leave latent traces of unlearned knowledge that can later re-emerge \citep{hu2024jogging,lucki2024adversarial,deeb2024unlearning,siddiqui2025dormant}.

\section{Data-Based Capability Control}

Since the capabilities of a foundation model are primarily driven by the diversity of its training dataset (assuming a sufficiently powerful learning algorithm) \citep{kaplan2020scaling,hoffmann2022chinchilla,brown2020gpt3}, the first level of control is the training distribution itself \citep{o2025deepignorance,maini2025safety}.
An updated model can be trained using standard training recipes on the intervened dataset---i.e., the one obtained via data-based interventions \citep{bourtoule2021shardedtrain}.

Note that data-based interventions rely on near-perfect identification recall---i.e., identifying (almost) all points that give rise to a particular capability---to ensure that the capability is fully removed \citep{cloud2024gradient,goel2024corrective,schoepf2025redirection}.
While conceptually straightforward, this approach shifts the challenge from controlling the model to attributing specific capabilities to specific subsets of the data distribution, which remains an open research problem \citep{grosse2023studying}.
Furthermore, filtering is complicated by the dual-use nature of foundational knowledge \citep{li2024wmdp} and the model's ability to re-derive harmful information from benign data \citep{shumailov2024ununlearning}.
Consequently, if unwanted capabilities persist after training, data-based methods necessitate computationally expensive retraining with an adjusted corpus \citep{alex2024protecting}.
These limitations motivate the exploration of \textit{learning-based approaches} for capability control, which present their own set of unique challenges, as discussed in the next section.

It is worth noting that these data-based interventions are applicable at all different phases of model training cycle, including pretraining and post-training.
Below, we consider three different data-based control mechanisms: (i) filtering the data distribution, (ii) curating the dataset carefully, and (iii) generating synthetic datasets.

\subsection{Data Filtering}

Dataset filtering attempts to collect the maximum number of datapoints initially.
Once the collection process has concluded, the aim is to remove datapoints that might contribute irrelevant capabilities or noise to the training process \citep{siddiqui2022metadata}.
Filtering is a common step in the pretraining dataset preparation for foundation models, where the focus is often more on the quantity of the dataset rather than the quality itself \citep{penedo2024fineweb}.

Recently, it has been argued that directly removing even the datapoints representing undesired capabilities can hamper the ability to exert control due to less developed representations in the model \citep{li2025bad}.
Similarly, augmenting data representing undesired capabilities such that it can lead to better control at later training stages is also a common strategy, and claimed to lead to better performance in many cases compared to just filtering \citep{maini2025safety,korbak2023pretraining}.
However, despite its challenges, data filtering has been argued to be an important ingredient for constructing tamper-resistant models \citep{o2025deepignorance}.

\subsection{Data Curation}

In contrast to filtering, where data is indiscriminately collected and subsequently pruned, data curation starts from a carefully selected set of capabilities that need to be distilled into the model.
The dataset is then collected such that it only gives rise to these desired capabilities \citep{michaud2025creation}.
In the case of dataset curation, the focus is on the quality of the dataset rather than just the quantity.
Hence, this is a common part of post-training pipelines, where only a small amount of high-quality data is desired to distill the right capabilities into the model \citep{alpaca,muennighoff2025s1}.

In practice, data curation cannot exhaustively cover all possible desirable/undesirable capabilities that we would like to encode into the model, and therefore, might be limited to the most general set of desired capabilities.

\subsection{Synthetic Data Generation}

Instead of filtering or curating a dataset, it is possible to generate synthetic data and train on the generated synthetic dataset \citep{eldan2023tinystories,alpaca,allal2025smollm2,sharma2025constitutional,cunningham2026constitutional}.
Synthetic dataset generation relies on carefully conditioning the generator to produce diverse and high-quality data \citep{allal2024cosmopedia,allal2025smollm2}.
This conditioning usually relies on careful seed inputs for the generation process, which are grounded in real data sources \citep{allal2024cosmopedia,alpaca}.
Failure to correctly condition the model can result in \textit{mode collapse} \citep{kirk2023understanding}, and consequently, \textit{model collapse} \citep{shumailov2024ai}.
The desired capabilities are indirectly encoded via the conditioning used to generate these datasets, which serves as a mechanism for data-based control \citep{sharma2025constitutional,cunningham2026constitutional}.

\section{Learning-Based Capability Control}

Learning-based approaches directly frame capability control as an optimization problem, which can then be solved using gradient-based approaches.
In contrast to data-based control, learning-based approaches often permit removal with imperfect identification, i.e., poor recall but high precision \citep{li2021antibackdoor,goel2024corrective,schoepf2025redirection}.

In most cases, learning-based control is applied after an initial round of pretraining and alignment (as highlighted in \Cref{fig:alignment-vs-cap}) \citep{rafailov2023dpo,zhang2024npo,li2024wmdp}.
This assumes that the model has already acquired both desirable and undesirable capabilities during pretraining.
It is worth noting that the approaches themselves are general and can be adapted to work directly at the time of pretraining \citep{korbak2023pretraining}.

We distinguish between the type of supervision used for model training, including behavior demonstrations, preference data, or direct unlearning, and the type of intervention used to control these capabilities, including direct weight updates, model editing, or representation engineering.

\subsection{Categorization based on Training Objective}

\subsubsection{Learning from Behavior Demonstrations}

One of the simplest and most straightforward learning-based approaches is to define the desired behavior when eliciting undesired capabilities, and hence, demonstrate the expected output behavior from the model.
In the context of language models, this refers to refusal training, where the model is trained to suppress responses to harmful queries by learning to refuse to comply with the specified user instructions \citep{andriushchenko2024does,yuan2025refuse}.

However, demonstrating the desired behavior in contexts that might elicit undesired capabilities requires not only a specification of the desired behavior but also the ability to identify the undesired behavior, which may be non-trivial in many cases.
Furthermore, behavior demonstrations are expected to generalize to relevant cases, which might lead to unexpected behaviors, such as over-refusal after refusal training \citep{cui2024or}.

\subsubsection{Learning from Preference Data}

The most fundamental component of any post-training pipeline for foundation models is learning from human preference data---i.e., Reinforcement Learning from Human Feedback (RLHF) \citep{christiano2017deep,rafailov2023dpo}.
RLHF attempts to train the model to conform to the specified user preferences \citep{ngo2022alignment,korbak2024aligning,edelman2025full}.
However, preferences obtained from humans are hard to scale, as each example produced by the model needs to be assigned an appropriate preference label.
Hence, RLHF relies on first training a reward model to capture the user's preferences \citep{abbeel2004apprenticeship,christiano2017deep}.
This reward model is then used to label data generated from the model on the fly, scaling the supervision appropriately.

There are recent variations of this paradigm, such as Reinforcement Learning from AI Feedback (RLAIF) \citep{lee2023rlaif} or Reinforcement Learning from Verifiable Rewards (RLVR) \citep{guo2025deepseek,zhao2025rlvr}, which rely on verifiable or machine-generated rewards instead of relying on humans for supervision.
Recent work has also explored strategies to directly optimize the model based on human preferences without explicitly training a reward model \citep{rafailov2023dpo}.

Training the model with RLHF reinforces behaviors preferred by the reward model (and therefore, implicitly the human raters), while simultaneously suppressing dispreferred behavior \citep{christiano2017deep}.
Note that while this implicitly acts as a control mechanism for undesired capabilities, the reward model particularly focuses on the behaviors demonstrated for training the reward model, and might not faithfully capture the right preferences for unexplored domains/situations.

RLHF-based capability control is also vulnerable to reward hacking, where the model over-optimizes the reward model by exploiting its weaknesses instead of truly learning to conform to the desired behavior \citep{ibarz2018reward,skalse2022defining}.
Furthermore, it only suppresses the undesired behavior at best (if both the reward model and the downstream policy are trained correctly), rather than removing it completely.
Therefore, advanced elicitation techniques can still elicit this behavior from the model \citep{che2025tamperingattacks}.

\subsubsection{Unlearning Undesired Behaviors}

Unlearning attempts to remove the influence of a specified set of datapoints to forget (commonly referred to as the \textit{forget set}), while still retaining information about the remaining datapoints (commonly referred to as the \textit{retain set}).
We distinguish between example-level unlearning methods primarily developed in the context of data privacy \citep{jia2023l1sparse,kurmanji2024scrub,chundawat2023badteacher,foster2024ssd,sepahvand2025selective,cooper2024machine}, from behavior unlearning methods particularly relevant in the context of safety of language models \citep{fan2023salun,zhang2024npo,li2024wmdp,zou2024circuitbreakers,tamirisa2408tamper,yao2024large}.
Behavior unlearning methods, using a small number of demonstrations of that behavior, attempt to completely unlearn the behavior rather than just altering the model outputs on the inputs provided for unlearning \citep{barez2025open}.
This makes it particularly difficult to both define and evaluate the success of unlearning algorithms in these cases.

Despite their promise, unlearning algorithms in many cases fail to correctly remove the influence of the unlearned datapoints or behaviors, resulting in either deliberate or accidental re-emergence \citep{qi2023finetuning,hu2024jogging,lucki2024adversarial,deeb2024unlearning,hayes2024inexact,siddiqui2025dormant}.
Such re-emergence has severe consequences, particularly when leveraging unlearning as a capability control mechanism.

\subsubsection{Adversarial Training}

Adversarial training formulates the task of elicitation of undesired capabilities as a min-max problem, where an adversary attempts to elicit these capabilities while the defender attempts to update the model to ensure that these capabilities cannot be elicited by the attacker \citep{goodfellow2014explaining,madry2017towards}.
This kind of min-max formulation has been used for defending language models against fine-tuning based attacks \citep{tamirisa2408tamper,che2025tamperingattacks}.

A defender needs to either remove the capability from the model for a powerful adversary, or make it hard for the adversary to elicit that capability by exploiting its weaknesses \citep{yuan2025llmadvunlearn}.
Fine-tuning-based adversaries can elicit dormant/suppressed capabilities \citep{li2024wmdp,che2025tamperingattacks,hu2024jogging}, and hence, are effective as inner adversaries to develop models resistant to these elicitation attacks \citep{tamirisa2408tamper}.

Aside from the prohibitive computational cost required for this approach to work, the robustness of the trained model is inherently bounded by the capabilities of the inner adversary used during training.
Consequently, adversarially trained models often fail to generalize to attacks not directly seen during training, remaining vulnerable to even minor differences in attack hyperparameters \citep{athalye2018obfuscated}.

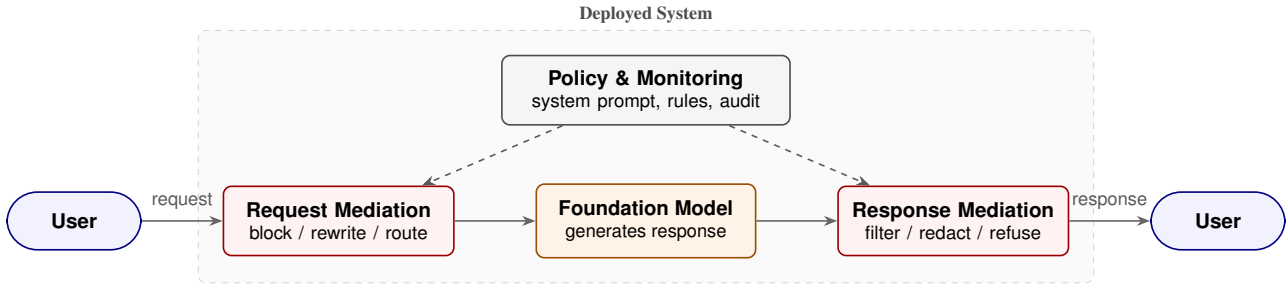
\begin{figure*}[t]
\centering
\resizebox{\textwidth}{!}{
\begin{tikzpicture}[
    font=\sffamily,
    node distance=1.1cm and 1.4cm,
    >=Stealth,
    core/.style={
        rectangle, rounded corners,
        draw=orange!60!black, fill=orange!10,
        thick, align=center, inner sep=8pt,
        minimum width=3.8cm, minimum height=1.2cm
    },
    gate/.style={
        rectangle, rounded corners,
        draw=red!60!black, fill=red!5,
        thick, align=center, inner sep=7pt,
        text width=3.5cm, minimum height=1.2cm
    },
    ctrl/.style={
        rectangle, rounded corners,
        draw=gray!60!black, fill=gray!8,
        thick, align=center, inner sep=7pt,
        text width=4.5cm, minimum height=1.1cm
    },
    user/.style={
        rectangle, rounded corners=15pt,
        draw=blue!50!black, fill=blue!5,
        thick, align=center, inner sep=6pt,
        text width=1.9cm, minimum height=1.0cm
    },
    flow/.style={thick, ->, color=gray!80!black},
    influence/.style={thick, dashed, ->, color=gray!70!black},
    label/.style={font=\footnotesize\sffamily, color=gray!70!black, midway, above=2pt, align=center}
]

\node[user] (user_in) {\textbf{User}};
\node[gate, right=of user_in] (req_gate)
{\textbf{Request Mediation}\\[-1pt]\footnotesize block / rewrite / route};

\node[core, right=of req_gate] (model)
{\textbf{Foundation Model}\\[-1pt]\footnotesize generates response};

\node[gate, right=of model] (resp_gate)
{\textbf{Response Mediation}\\[-1pt]\footnotesize filter / redact / refuse};

\node[user, right=of resp_gate] (user_out) {\textbf{User}};

\draw[flow] (user_in) -- node[label]{request} (req_gate);
\draw[flow] (req_gate) -- (model);
\draw[flow] (model) -- (resp_gate);
\draw[flow] (resp_gate) -- node[label]{response} (user_out);

\node[ctrl, above=1.0cm of model] (policy)
{\textbf{Policy \& Monitoring}\\[-1pt]\footnotesize system prompt, rules, audit};

\draw[influence] (policy) -- (req_gate);
\draw[influence] (policy) -- (resp_gate);

\begin{scope}[on background layer]
\node[fit=(req_gate)(model)(resp_gate)(policy),
      draw=gray!35, dashed, rounded corners, inner sep=12pt, fill=gray!4] (boundary) {};
\node[above, font=\bfseries\footnotesize, color=gray!60!black] at (boundary.north)
{Deployed System};
\end{scope}

\end{tikzpicture}
}
\caption{
\textbf{System-Based Capability Control.} This layer enforces control at inference time via wrappers on the system's inputs and outputs. These mechanisms can intercept, block, redact, or route traffic based on safety policies and monitoring signals, operating independently of the model's internal weights and representations.
}
\label{fig:system-control}
\end{figure*}

\subsection{Categorization based on Mechanism of Intervention}

We consider learning-based interventions here, including both training-time interventions that permanently change the model, and inference-time interventions that only steer model activations.
Training-time interventions can suppress or completely remove undesired capabilities \citep{tamirisa2408tamper}, while inference-time interventions are only capable of suppression \citep{zou2023representation}.

\subsubsection{Weight Updates to the Model}

The most obvious intervention is to just fine-tune the weights of the model in order to make the elicitation of undesired capabilities difficult (in the case of suppression) or even impossible (in the case of complete removal) \citep{tamirisa2408tamper,kurmanji2024scrub,christiano2017deep}.
This also covers training of low-rank adapters instead of all the weights of the model \citep{hu2022lora,liu2024dora}.

\subsubsection{Locating and Editing Knowledge}

Instead of modifying all the weights of the model via fine-tuning, an alternate strategy is to locate parameters within the model responsible for a particular prediction using tools from causal analysis, and directly edit those parameters in order to adjust the model's prediction \citep{santurkar2021editing,meng2022locating,meng2022mass,shah2024decomposing,zhang2024comprehensive,wang2024knowledge}.
However, such interventions are local in nature, and hence, fail to generalize in many cases \citep{hsueh2024editing,liu2026we}.

As these techniques are primarily used for editing specific predictions from the model, it is unclear how these interventions generalize to the larger capability control landscape, where complete capabilities need to be removed from the model.
Note that some selective editing strategies also resemble selective weight updates prevalent in the unlearning literature \citep{foster2024ssd,anthropic_selective_gradient_masking_2025,torkzadehmahani2024improved}.

\subsubsection{Representation Engineering}

A dominant class of inference-time interventions is based on activation steering, where the activations of a model are steered to either amplify or suppress different capabilities \citep{turner2023steering,li2023inference,zou2023representation}.
Note that while the intervention is applied at inference time, there is still dependence on a dataset as well as some training-time computation to find the appropriate steering vector \citep{arditi2024refusal}.
We consider interventions applied via the use of Sparse Auto-Encoders (SAEs) as instantiations of this framework \citep{cunningham2023sparse,gao2024scaling}.

These interventions can only suppress model capabilities as they do not directly alter the weights of the model.
Furthermore, most of these interventions are argued to be weaker in terms of their effectiveness compared to simple baselines based on prompting \citep{wu2025axbench}.

\section{System-Based Capability Control}

It is difficult to provide deterministic guarantees when relying solely on data-based or learning-based mechanisms, as they are inherently probabilistic.
This lack of guarantees is unacceptable for high-stakes scenarios, particularly in the integration of sensitive tools (e.g., banking APIs) within agentic systems.
In such cases, it is catastrophic if the model accesses a tool inappropriately, regardless of the point of origin of this failure \citep{greshake2023youve,liu2023prompt,wu2024systemleveldefenseindirectprompt,costa2025securing}.
Consequently, we identify system-based capability control as a primary line of defense for future agentic systems, even offering deterministic guarantees in some cases for models that take direct actions in the real world \citep{yao2022react,costa2025securing,siddiqui2024permissive}.

These policies provide stronger assurances by design, independent of the model's internal weights or training distribution.
However, it is worth noting that even system-level defenses---particularly those relying on prompting or monitoring the model inputs or outputs---may be circumvented due to the failure of the model to follow instructions, failure of the monitoring classifiers, or due to the employment of more sophisticated behaviors such as (internal) alignment faking, where models conceal their true intent or reasoning processes \citep{greenblatt2024alignment,korbak2025cotmonitor}.
Such guardrails can also be directly bypassed when considering open-weight model releases \citep{tamirisa2408tamper}.
Finally, system-based interventions can significantly increase end-to-end latency because they operate at inference time \citep{costa2025securing}, unlike the predominantly training-time techniques discussed above.

The dominant approaches recently explored at the system level can be broadly categorized as: (i) system prompts, (ii) input/output filtering, (iii) chain-of-thought monitoring, and (iv) information-flow analysis.
A schematic overview of system-based control approaches is presented in \Cref{fig:system-control}.

\subsection{System Prompts}

Given the capabilities of recent foundation models to learn within context \citep{lu2023emergent}, prompting has been used in a range of ways to steer model behavior, including unlearning \citep{bhaila2025soft,pawelczyk2023context,thaker2024guardrail}, as well as defining model policy directly via natural language \citep{bai2022constitutional,palla2025policy,claudeconstitution}.

While effective in many cases and an essential component of all frontier model deployments, such prompt-based guardrails---most commonly known as \textit{system prompts} \citep{mu2025closer}---are easy to bypass via prompt injection \citep{liu2023prompt} due to the difficulty in enforcing instruction hierarchy in practice \citep{wallace2024instruction}.
Furthermore, they fail to provide any deterministic assurances regarding model behavior at inference time.

\subsection{Input/Output Filtering}

One of the most prevalent approaches to enforce capability control at the system level is by filtering inputs and outputs using separate classifier models \citep{sharma2025constitutional,cunningham2026constitutional}.
These mechanisms are a standard component of modern production pipelines, commonly instantiated as \textit{moderation APIs} or guardrails \citep{openai_moderation_guide,anthropic_moderation_guide,han2024wildguard,inan2023llamaguard}.
In contrast to data-based control, this filtering is applied at inference time, where certain inputs are rejected, or outputs are redacted by the system---i.e., classifiers intercept potentially harmful content before it reaches the foundation model (input filtering) or the user (output filtering).
However, note that the defense mechanism provided by such systems is again probabilistic in nature due to the probabilistic nature of the classifiers \citep{han2024wildguard,inan2023llamaguard}.

\subsection{Chain-of-Thought Monitoring}

Chain-of-Thought (CoT) \citep{wei2022chain,kojima2022zeroshotcot}, where the model is asked to reason through the intermediate steps before producing a final response, has been a major driving force for improving the performance of language models, including the recently introduced test-time scaling paradigm \citep{guo2025deepseek,muennighoff2025s1}.
As these models think via writing their outputs to an internal thinking scratch-pad (which is not directly accessible by end-users in many cases), it is possible to catch problematic model behaviors by monitoring the chain-of-thought traces from these models \citep{korbak2025cotmonitor}.
The output of these monitors can then be used to control the output of the model, such as suppression of the model output via moderation APIs (which can in turn be similar to the output classifiers defined in the input/output filtering approach) \citep{han2024wildguard}.

While this is fragile \citep{barez2025chain,chen2025reasoning}, it is argued that these traces are much more faithful to the internal reasoning of the model when used as a way to scale test-time compute in order to solve complex problems, compared to rationalizing its decisions in a post-hoc fashion \citep{emmons2025cotmonitor}.

\subsection{Information-Flow Control}

As the provided in-context information flows through the language models, it is possible to leverage ideas from information-flow analysis to understand the influence of each of the provided in-context examples on the model's outputs \citep{wutschitz2023rethinking,siddiqui2024permissive,costa2025securing}.
Such techniques can then be used to enforce information-flow control in order to guarantee the safety of the overall system \citep{clause2007dytan}.

Considering scenarios where each provided auxiliary piece of information can be assigned an appropriate reliability, security, or accessibility label, it is possible to estimate the reliability, safety, or accessibility label of the model output based on taint tracking \citep{wutschitz2023rethinking,siddiqui2024permissive,costa2025securing}.
Appropriate access control policies on model capabilities can then be defined on the basis of these labels \citep{siddiqui2024permissive,costa2025securing,lan2025contextual}.
As an example, access to sensitive capabilities such as operating a bank account can be restricted in such cases \citep{greshake2023youve,costa2025securing}.

While promising as it can provide deterministic guarantees, the assumption regarding access to appropriate and fully trusted labels can be difficult to satisfy in practice.
Furthermore, estimating the influence of the provided in-context examples can even exceed the cost of the response generation itself, making it economically infeasible to apply in practice.

\section{Open Challenges and Directions}
\label{sec:open_challenges}

\subsection{Evaluating the Effectiveness of Control Mechanisms}

Evaluating the effectiveness of a control mechanism is notoriously difficult, primarily due to the adversarial dynamics between the attacker and the controller.
Such difficulties have plagued related domains such as adversarial robustness \citep{carlini2017towards} and approximate unlearning \citep{che2025tamperingattacks,li2024wmdp,hu2024jogging,triantafillou2024unlearningcomp}.
The core challenge in capability control is distinguishing between \textit{suppression} (hiding a capability) and \textit{removal} (erasing it).
Current benchmarks often fail to detect latent capabilities that can be re-elicited via fine-tuning or jailbreaking \citep{zou2023universal,qi2023finetuning,hu2024jogging,siddiqui2025dormant}.
Developing rigorous evaluations that can certify the \textit{absence} of a capability---rather than just its failure to manifest---remains a critical open challenge.

\subsection{Open-Weight Models}

Enforcing robust system-level control in open-weight models is inherently challenging, as the user can modify or bypass any component of the system pipeline.
Learning-based control is similarly difficult to maintain in these settings, as safety guardrails can be eroded upon fine-tuning \citep{che2025tamperingattacks,tamirisa2408tamper,qi2023finetuning}.
Therefore, we posit that the most robust line of defense for open-weight models is controlling the training distribution---a strategy that has recently been employed to achieve robustness even with white-box access to the model \citep{o2025deepignorance,maini2025safety}.

\subsection{Dual-Use Capabilities}

Most fundamental capabilities are inherently dual-use.
For example, knowledge of virology is essential for vaccine development, yet the same knowledge can facilitate the synthesis of pathogens \citep{li2024wmdp}.
Consequently, it is difficult to define absolute guardrails for removing harmful capabilities without compromising the core knowledge required for benign tasks.
This creates a fundamental conflict between safety and utility: to maintain a capable model, we are often forced to retain knowledge that carries inherent risks.

\subsection{Compositional Generalization}

Even if specific harmful knowledge is successfully removed, large models may be able to resynthesize it by composing remaining fundamental concepts.
For instance, a model might recover a dangerous capability via in-context learning by combining benign facts or reasoning steps that were left intact \citep{shumailov2024ununlearning}.
This compositional nature of intelligence makes it inherently difficult to provide absolute guarantees against the re-emergence of capabilities, as the model can effectively rediscover harmful knowledge using its general reasoning abilities.

\section{Call for Research: \textit{Defense-in-Depth} Approach}
\label{sec:defense_in_depth}

This paper has analyzed data, learning, and system controls as distinct layers for capability control.
Data-based controls act as the first line of defense but suffer from imperfect recall \citep{o2025deepignorance}.
Learning-based control internalizes safety, but is reversible via parameter updates \citep{che2025tamperingattacks}.
System-based approaches can provide deterministic guarantees, but carry the risk of being directly bypassed for open-weight releases \citep{han2024wildguard,costa2025securing}.

As highlighted earlier, distinct use cases are best addressed with different types of mechanisms.
For example, data-level control is paramount for open-weight models, whereas system-level control is best suited to guard against harmful generations arising from compositional generalization.
Ultimately, no single mechanism is effective across all scenarios.
Therefore, we argue that \textit{effective capability control requires defense-in-depth across data, learning, and system layers, compensating for the inherent weaknesses of each individual layer}.

\section{Alternate Views}

While we argue for capability control as a distinct goal, we address two prominent counter-arguments.

The first is the ``Alignment Sufficiency'' hypothesis, which posits that perfect instruction-following renders capability control redundant; a fully aligned model should naturally exert control over harmful capabilities \citep{askell2021general}.
However, as discussed in \Cref{subsec:diff_alignment_cap}, alignment is inherently context-dependent, which enables unique attack vectors via the careful reframing of user prompts.
Consequently, we argue that relying on context-sensitive alignment is insufficient; security requires absolute prohibitions on specific capabilities to serve as a hard backstop against alignment failures \citep{WEF_2025_AI_redlines,claudeconstitution}.
Moreover, given the known trade-offs between alignment and utility, commonly referred to as the \textit{alignment tax} \citep{alignment_forum_alignment_tax,lin2024alignmenttax} (see \Cref{subsec:diff_alignment_cap}), we argue that leveraging the full capability control stack mitigates these costs. Placing the entire burden on post-training alignment—which relies on modifying model weights—risks greater performance degradation than a distributed, multi-layer approach.
Collectively, these arguments demonstrate that, while alignment can be a useful ingredient within a capability control framework, adopting the ``Alignment Sufficiency'' view is at best suboptimal, and at worst catastrophic.

The second objection concerns the ``Futility of Removal,'' which suggests that knowledge in large models is so diffuse and interconnected that removing specific hazardous capabilities without degrading general reasoning is practically impossible \citep{li2024wmdp}.
While we acknowledge the difficulty of perfect removal, we contend that \textit{raising the cost of elicitation} is a valuable security goal in itself.
Even if removal is insufficient to defend against a resource-unbounded adversary with unrestricted access to model weights and training data, making progress on this front significantly reduces the threat from the vast majority of realistic, resource-constrained adversaries.
We advocate for a defense-in-depth approach (\Cref{sec:defense_in_depth}) precisely because the interventions at any single layer are imperfect.

We argue that by combining several capability control layers and encouraging future work on synergistic mechanisms, we can build systems that are robust against a much wider range of realistic threat models.

\section{Conclusion}

This paper argues for distinguishing the goal of \textit{capability control} from that of \textit{alignment}.
We contend that alignment's inherent context-dependence makes it an insufficient guardrail for high-stakes capabilities, and that conflating the two risks exacerbating the trade-off between safety and model utility.
We structured the landscape of control into three layers: (i) the training distribution, (ii) the model's parameters and representations, and (iii) system-level constraints.
Each layer possesses inherent vulnerabilities and distinct trade-offs regarding computational efficiency, data access, and deterministic guarantees.
Crucially, while specific layers may be better suited to address specific types of harms, no single layer is sufficient in isolation.
We therefore advocate for a \textit{defense-in-depth} approach where individual layers work synergistically to complement each other's strengths and mitigate individual weaknesses.
We invite the research community to adopt this framework to advance the development of robust, safe, and secure foundation models.

\section*{Acknowledgements}

The authors would like to thank Michael C. Mozer for useful comments on the initial draft of this work.
AW acknowledges support from Turing AI Fellowship under grant EP/V025279/1, the Alan Turing Institute, and the Leverhulme Trust via CFI.

\bibliography{main}
\bibliographystyle{icml2026}

\end{document}